# Of Starships and Klingons: Bayesian Logic for the 23$^{rd}$ Century

**Kathryn B. Laskey and Paulo C. G. da Costa**

George Mason University
4400 University Drive
Fairfax, VA 22030-4400
[pcosta, klaskey]@gmu.edu

## Abstract

Intelligent systems in an open world must reason about many interacting entities related to each other in diverse ways and having uncertain features and relationships. Traditional probabilistic languages lack the expressive power to handle relational domains. Classical first-order logic is sufficiently expressive, but lacks a coherent plausible reasoning capability. Recent years have seen the emergence of a variety of approaches to integrating first-order logic, probability, and machine learning. This paper presents Multi-entity Bayesian networks (MEBN), a formal system that integrates First Order Logic (FOL) with Bayesian probability theory. MEBN extends ordinary Bayesian networks to allow representation of graphical models with repeated sub-structures, and can express a probability distribution over models of any consistent, finitely axiomatizable first-order theory. We present the logic using an example inspired by the Paramount Series *Star Trek*.

## 1 INTRODUCTION

Uncertainty is a ubiquitous feature of the world, and probability theory is a natural candidate to represent uncertain phenomena. Bayesian networks (BNs) enable parsimonious specification and tractable inference for realistically complex probability distributions, and have been applied to a wide variety of domains (Heckerman *et al.* 1995). However, the simple attribute-value representation of BNs is insufficiently expressive for relational domains – that is, domains in which many entities of different types interact with each other in varied ways. Relational domains can be represented using first-order logic (FOL), which has become the *de facto* standard for logical systems from both a theoretical and practical standpoint. However, systems based on classical first-order logic lack a widely accepted and logically coherent methodology for reasoning under uncertainty.

A number of languages have appeared that extend the expressiveness of standard BNs to relational domains. Characterizing the formal properties of such languages is an active area of research. This paper discusses some of the primary representational challenges that must be addressed by a logical formalism that combines first-order logic and probability. Our vehicle for presenting these ideas is Multi-entity Bayesian networks (MEBN), a knowledge representation formalism that combines first-order logic with Bayesian probability (Laskey 2005). MEBN syntax is designed to highlight the relationship between a MEBN theory and its FOL counterpart. Although our examples are presented using MEBN, our main focus is on logical concepts that are to a large degree independent of the particular syntax used to express a first-order probabilistic domain theory.

MEBN is a full first-order Bayesian logic. Its syntax, model construction and inference processes, and semantics provide a means of defining, querying, and interpreting probability distributions over unbounded and possibly infinite numbers of interrelated hypotheses. MEBN can express a joint distribution over models of any consistent, finitely axiomatizable first-order theory, and can add new axioms via Bayesian conditioning (Laskey, 2005). The following sections use a running example to motivate the need for more expressive probabilistic representations and to describe how MEBN meets the need. Results are stated on the expressive power of MEBN. Related work on integrating logic and probability is discussed.

## 2 ON PLANETS AND STARSHIPS

The setting for our case study is the Starship *Enterprise* in the late 23$^{rd}$ Century. Our task is to detect Klingon starships (our enemies) and assess the level of danger they pose to our own starship, the *Enterprise*. Starship detection is performed by the *Enterprise*'s suite of sensors. However, Klingon starships may be in "cloak mode," which makes them invisible to the *Enterprise*'s sensors. The only hint of a nearby starship in cloak mode is a slight magnetic disturbance caused by the enormous amount of energy required for cloaking. The *Enterprise* has a magnetic disturbance sensor, but it is very hard to distinguish background magnetic disturbance from that generated by a nearby starship in cloak mode.

We could use a BN to model features of an approaching starship – e.g., its type, potential to harm the *Enterprise*, position, etc. In addition, we could infer intention from these characteristics and our previous knowledge of similar situations. However, there are many questions of interest to the *Enterprise* and its crew that demand greater expressive power than standard BNs can offer. As an example, we cannot know in advance how many starships the *Enterprise* is going to encounter. Even if we were to build a BN for each possible number of nearby starships, if the number of nearby starships is uncertain, we would not know which one to use. We also cannot specify in advance the relationships among nearby ships, e.g., whether they are isolated ships operating independently or are acting as a group in concert. In short, BNs lack the expressive power to represent entity types (e.g., starships) that can be instantiated as many times as required for the situation at hand, and can be related to each other in varied ways (e.g., operate in groups).

Another known limitation of BNs is their lack of support for recursion. For example, the magnetic disturbance caused by a starship in cloak mode would show a characteristic temporal pattern. Standard BNs do not provide a natural way to represent such repeated patterns. Dynamic Bayesian networks (DBNs) (Murphy 1998) and partially dynamic Bayesian networks (e.g. Takikawa *et al.* 2001) extend BNs to model temporal patterns. However, there is no standard means to represent general recursive probabilistic relationships.

This section has provided just a glimpse of the issues that must be confronted when applying Bayesian networks to complex problems. The next section extends our model to show how MEBN logic handles many of the difficulties commonly encountered in knowledge representation.

## 3 USING MEBN LOGIC

Like present-day Earth, 23rd Century outer space is not a politically trivial environment. Thus, we are likely to encounter different alien species with diverse profiles. Although MEBN logic can represent the full range of species inhabiting the Universe in that epoch, for this paper we limit the explicitly modeled species to Friends, Klingons, Romulans, and the catch-all category *Unknown*. A truly "realistic" model would also consider each starship's type, offensive power, ability to inflict harm to the *Enterprise* given its range, and other features pertinent to the model's purpose. We will add some of these features to our model as we present the basic constructs of MEBN logic, and will argue that MEBN provides sufficient expressive power for arbitrarily complex domain theories.

### 3.1 UNDERSTANDING MFRAGS

MEBN logic represents the world as comprised of entities that have attributes and are related to other entities. Random variables (RVs) represent features of entities and relationships among entities. Knowledge about attributes and relationships is expressed as a collection of MEBN fragments (MFrags) organized into MEBN Theories (MTheories). An MFrag represents a conditional probability distribution for instances of its resident RVs given their parents in the fragment graph and the context RVs. An MTheory is a set of MFrags that collectively satisfies consistency constraints ensuring the existence of a unique joint probability distribution over instances of the RVs represented in each of the MFrags within the set.

Like a BN, an MFrag contains nodes arranged in a directed graph. Nodes represent RVs; arcs represent direct dependency relationships; and local distributions specify conditional probability distributions for resident nodes. Each node has an associated RV label and a parameterized list of arguments. Entity identifiers are substituted for arguments to form instances of the RVs.

For example, the MFrag of Figure 1 represents knowledge about the degree of danger to which our own starship is exposed. The fragment graph has eight nodes. The five nodes at the top of the figure are context nodes; the two darker nodes below the context nodes are input nodes; and the remaining node, labeled *HarmPotential*($st, t$), is a resident node. The arguments *st* and *t* are placeholders for a potentially harmful entity and a time step, respectively. To refer to an actual entity, unique identifiers are substituted for the arguments. When no confusion is likely to result, the term RV is used both for the template and for the instances themselves. By convention, unique identifiers begin with an exclamation point, and no two distinct entities may have the same unique identifier. For example, *HarmPotential*(!$ST$1, !$T$1) and *HarmPotential*(!$ST$2, !$T$1) are two instances of the RV template *HarmPotential*($st, t$) that both occur at time step !$T$1.

The resident nodes of an MFrag have local distributions that define probabilities for instances of the node given instances of the node's parents in the fragment graph. In a complete MTheory, every node of any MFrag has exactly one *home MFrag*, where its local distribution is defined. Input and context nodes (e.g., *OpSpec*($st$) or *IsOwnStarship*($s$)) influence the distribution of the resident nodes, but their distributions are defined in their home MFrags.

Context nodes represent conditions that must be satisfied for the influences and local distributions of the fragment graph to apply. Context nodes are Boolean, with possible values *True*, *False*, and *Absurd*.[1] As an example, if the *Enterprise* unique identifier !$ST$0 is substituted for *s* in *IsOwnStarship*($s$), the resulting hypothesis is true, and the context node is satisfied. Substituting a different starship unique identifier (say, !$ST$1) for *s* makes the hypothesis false. Finally, substituting the unique identifier of a non-starship (say, !$Z$1), makes the result absurd (i.e., it is absurd to ask whether or not a zone in space is one's own starship).

---

[1] The symbols T, F, and ⊥ are used in place of *True*, *False* and *Absurd* in Laskey (2005).

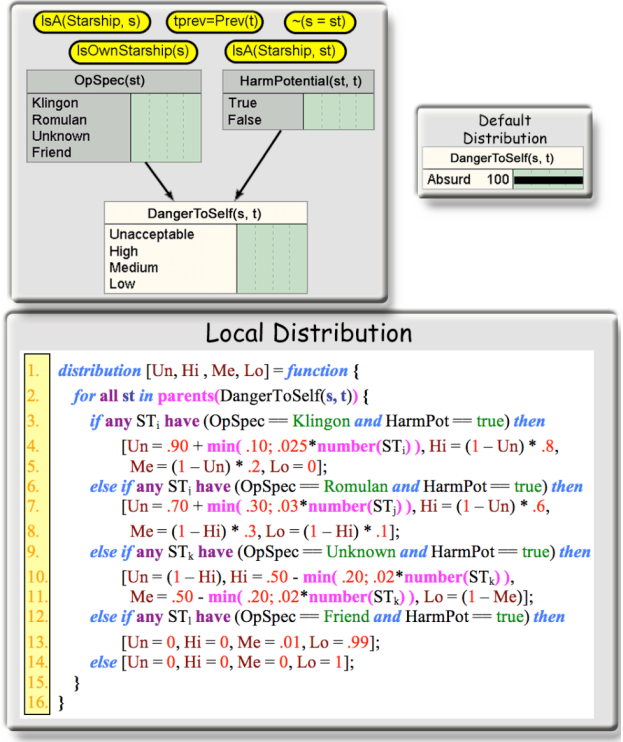

Figure 1 – DangerToSelf MFrag

**Definition 1** (from Laskey 2005): An *MFrag* $\mathcal{F} = (C, \mathcal{I}, \mathcal{R}, \mathcal{G}, \mathcal{D})$ consists of a finite set $C$ of *context* RVs; a finite set $\mathcal{I}$ of *input* RVs; a finite set $\mathcal{R}$ of *resident* RVs; a *fragment graph* $\mathcal{G}$; and a set $\mathcal{D}$ of *local distributions*, one for each member of $\mathcal{R}$. The sets $C$, $\mathcal{I}$, and $\mathcal{R}$ are pairwise disjoint. The fragment graph $\mathcal{G}$ is an acyclic directed graph whose nodes are in one-to-one correspondence with the RVs in $\mathcal{I} \cup \mathcal{R}$, such that RVs in $\mathcal{I}$ correspond to root nodes in $\mathcal{G}$. Local distributions specify conditional probability distributions for the resident RVs as described in Definition 2 below. ∎

Local distributions in standard BNs are typically specified as static tables. This limits each node to a fixed number of parents. On the other hand, an instance of a node in an MTheory might have any number of parents (even infinitely many!), and a node's local distribution must specify how to combine these influences. Thus, MEBN implementations (i.e. languages based on MEBN logic) must provide an expressive language for defining local distributions. The local distribution of Figure 1 uses pseudo-code to convey the idea of using local expressions to specify probability distributions, while not committing to a particular syntax. It can be seen from this local expression that the most uncomfortable situation for the *Enterprise* occurs when there are many Klingons with potential to harm the *Enterprise*; fewer potentially harmful Klingons means less danger; the danger is still less if the only potentially harmful nearby ships are Romulans; and the best situation occurs when there are no nearby ships with potential to harm the *Enterprise*.

The set of all instances of the parents of a resident node and the context nodes of its respective home MFrag is called the *partial world* for that resident node. A *partial world state* for a given partial world is an assignment of values to each RV in the partial world. A configuration of parent and context RVs within a partial world can influence the node's distribution only if the context constraints are satisfied. The *influence counts* (Laskey 2005) tally the number occurrences in a partial world of each configuration of parent and context RVs such that the context constraints are satisfied. In MEBN logic, the local distribution for a node is a function of the influence counts.

**Definition 2** (from Laskey 2005): The *local distribution* $\pi_\psi$ for resident RV $\psi$ in MFrag $\mathcal{F}$ is a function $\pi_\psi(\alpha|S)$ that maps unique identifiers $\alpha$ and partial world states $S$ to real numbers, such that the following conditions are satisfied:

- 2a. For a given partial world state $S$, $\pi_\psi(\alpha|S)$ is a probability distribution on the unique identifier symbols. That is, $\pi_\psi(\alpha|S) \geq 0$ and $\Sigma_\alpha \pi_\psi(\alpha|S) = 1$, where $\alpha$ ranges over the unique identifier symbols.
- 2b. There is a recursively enumerable subset $\{\alpha_{\psi 1}, \alpha_{\psi 2}, \ldots\}$ of unique identifier symbols, called the *possible values* of $\psi$, s.t. $\pi_\psi(\{\alpha_{\psi 1}, \alpha_{\psi 2}, \ldots\} | S) = 1$ for all $S$.
- 2c. There is an algorithm s.t. for any recursive subset $A$ of the possible values of $\psi$ and any partial world state $S$ for $\psi$, either the algorithm halts with output $\pi_\psi(A|S)$ or there exists a value $N(A,S)$ s.t. if the algorithm is interrupted after a number of time steps greater than $N(A,S)$, the output is $\pi_\psi(A|S)$.
- 2d. $\pi_\psi$ depends on the partial world state only through the influence counts. That is, any two partial world states having the same influence counts map to the same probability distribution;
- 2e. Let $S_1 \subset S_2 \subset \ldots$ be an increasing sequence of partial world states for $\psi$. There exists an integer $N$ such that if $k > N$, $\pi_\psi(S_k) = \pi_\psi(S_N)$.

The probability distribution $\pi_\psi(\alpha|\varnothing)$ is called the *default distribution* for $\psi$. It is the probability distribution for $\psi$ given that no potential influencing configurations satisfy the conditioning constraints of $\mathcal{F}$. If $\psi$ is a root node in an MFrag $\mathcal{F}$ containing no context constraints, then the local distribution for $\psi$ is just the default distribution. ∎

Continuing with the example of Figure 1, to find the probability distribution for an instance of *DangerToSelf*(*s, t*), we first identify all instances of *HarmPotential*(*st, t*) and *OpSpec*(*st*) for which the context constraints are satisfied. If there are no such instances, then the default distribution specifies that *DangerToSelf*(*s, t*) has value *Absurd* with probability 1; otherwise, we use the MFrag's local distribution for its resident node, *DangerToSelf*(*s, t*).

Clearly, we could have included additional detail and explored many nuances. For example, we could have specified a local distribution in which a large number of nearby Romulan ships poses greater danger than an isolated Klingon ship, because it might indicate a coordinated Romulan attack. The example was purposely kept simple in order to clarify the basic capabilities of the logic. Yet, it is clear that more complex knowledge patterns could be accommodated as needed to suit the requirements of the application. MEBN logic has built-in logical MFrags for logical connectives, function composition, and quantifiers. These logical MFrags provide the ability to express any first-order sentence. Thus, MEBN has sufficient expressive power to represent virtually any scientific hypothesis.

MEBN also provides theoretically grounded support for representing very general forms of recursion via MFrags that allow influences between instances of the same RV template. Allowable recursive definitions must ensure that no RV instance can influence its own probability distribution. As in non-recursive MFrags, the input nodes in a recursive MFrag may include nodes whose local distributions are defined in another MFrag. In addition, the input nodes may include instances of recursively-defined nodes in the MFrag itself.

MFrags provide a flexible means to represent knowledge about specific subjects within the domain of discourse, but the true gain in expressive power is revealed when we aggregate these "knowledge patterns" to form a coherent domain theory that can be applied to reason about specific situations and refined through learning. The following section describes how to combine MFrags to form coherent theories of a domain.

### 3.2 BUILDING MEBN THEORIES

In order to build a coherent model we have to make sure that our set of MFrags collectively satisfies consistency constraints ensuring the existence of a unique joint probability distribution over instances of the RVs mentioned in the MFrags. Such a coherent collection of MFrags is called an MTheory. For example, an MTheory cannot have multiple conflicting distributions for a RV in different MFrags, or sets of MFrags with cyclic influences. Some of the consistency conditions of Definition 3 below involve the notion of an *ancestor chain* for a RV instance, which is a sequence of RV instances, each of which influences the next RV instance in the chain, terminating in the given RV instance.

***Definition 3*** (from Laskey 2005): Let $\mathcal{T} = \{\mathcal{F}_1, \mathcal{F}_2 \dots\}$ be a set of MFrags. Let $\mathcal{V}_\mathcal{T}$ denote the set of RV terms contained in the $\mathcal{F}_i$, and let $\mathcal{N}_\mathcal{T}$ denote the set of RV instances that can be formed from $\mathcal{V}_\mathcal{T}$. $\mathcal{T}$ is a *simple MTheory* if the following conditions hold:

- 3a. *No cycles*. No RV instance is an ancestor of itself;
- 3b. *Bounded causal depth*. For any instance $\phi(\alpha) \in \mathcal{N}_\mathcal{T}$, there exists an integer $N_{\phi(\alpha)}$ s.t. if $\phi_d(\alpha_d) \to \phi_{d-1}(\alpha_{d-1})$ $\to \dots \to \phi(\alpha)$ is an ancestor chain for $\mathcal{T}$, then $d \leq N_{\phi(\alpha)}$. The smallest such $N_{\phi(\alpha)}$ is called the *depth* $d_{\phi(\alpha)}$ of $\phi(\alpha)$.
- 3c. *Unique home MFrags*. For each $\phi(\alpha) \in \mathcal{N}_\mathcal{T}$, there exists exactly one MFrag $\mathcal{F}_{\phi(\alpha)} \in \mathcal{T}$, called the *home MFrag* of $\phi(\alpha)$, such that $\phi(\alpha)$ is an instance of a resident RV $\phi(\theta)$ of $\mathcal{F}_{\phi(\alpha)}$.
- 3d. *Recursive specification*. $\mathcal{T}$ may contain infinitely many domain-specific MFrags, but if so, the MFrag specifications must be recursively enumerable. That is, there must be an algorithm that lists a specification (input, output, and context RVs, fragment graph, and local distributions) for each MFrag in turn, and eventually lists a specification for each MFrag. ∎

***Theorem 1:*** Let $\mathcal{T} = \{\mathcal{F}_1, \mathcal{F}_2 \dots\}$ be a simple MTheory. There is a joint probability distribution $\mathcal{P}_\mathcal{T}^{\text{gen}}$ on the set of instances of the RVs in its MFrags that is consistent with the local distributions assigned by the MFrags of $\mathcal{T}$. ∎

***Proof sketch:*** The proof is by induction on the maximum depth of ancestor chains. Assuming as an induction hypothesis that mutually consistent joint distributions exist for all finite sets of RV instances containing only instances of depth no greater than $d$, Kolmogorov's existence theorem (Billingsley 1995) implies the existence of a joint distribution on the set of all RV instances of depth no greater than $d$. From this, the induction hypothesis follows for depth $d+1$. A second application of Kolmogorov's existence theorem implies existence of a joint distribution on all random variable instances. For a full proof, see Laskey (2005). ∎

A *generative* MTheory summarizes statistical regularities that characterize a domain. These regularities are captured and encoded in a knowledge base using a combination of expert judgment and learning from observation. To apply a generative MTheory to reason about particular scenarios, we need to provide specific information about the individual entity instances involved in the scenario. Given this information, Bayesian inference can answer specific questions of interest (e.g., how great is the current level of danger to the *Enterprise*?), and also refine the MTheory (e.g., each encounter with a previously unknown species provides additional data about the level of danger to the *Enterprise* from a starship operated by the species). Bayesian inference is used to perform both problem-specific inference and learning. *Findings* are the basic mechanism for incorporating observations into MTheories. A finding is represented as a special 2-node MFrag containing a node from the generative MTheory and a node declaring one of its states to have a given value. From a logical point of view, inserting a finding into an MTheory corresponds to asserting a new axiom in a first-order theory. In other words, MEBN logic is inherently open, having the ability to incorporate new axioms as evidence and update the probabilities of all RVs.

*Finding absorption* restructures an MTheory with findings into an equivalent generative MTheory.

Any consistent, finitely axiomatizable FOL theory can be translated into an infinity of MTheories, all having the same purely logical consequences, but assigning different probabilities to statements whose truth-value is not determined by the axioms of the FOL theory. MEBN logic contains a set of built-in logical MFrags, including logical connective, function composition and quantifier MFrags, that provide the ability to represent any sentence in first-order logic. If additional conditions are satisfied, a conditional distribution exists given any finite sequence of findings that does not contradict the generative MTheory (Laskey 2005). Thus, MEBN logic can express a joint distribution over models of any finitely axiomatizable FOL theory. MEBN semantics is consistent with the definition from mathematical statistics of random variables as measurable functions from a probability space to an outcome space. MEBN logic integrates Bayesian probability and statistics with first-order logic, and provides a logical foundation for open-world systems that incorporate evidence in a logically coherent manner.

Figure 2 shows an example of a generative MTheory for the *Star Trek* scenario. For the sake of brevity, local distribution formulas and default distributions are not shown. The *Entity Type* MFrag declares the possible types of entity represented by the MTheory. The other MFrags encode knowledge about attributes and behavior of entities of each type, and of their relationships to each other.

FOL provides a theoretical foundation for the type systems used in popular object-oriented and relational languages. A typed version of MEBN logic extends FOL-based type systems to include probability. Our simple example represents only four entity types. The parent of *Type*($e$) is the *identity* RV $\Diamond(e)$, which maps a unique identifier either to itself or to *Absurd*. There is a set of reserved unique identifiers for each type. There may be no upper bound on the number of identifiers for a given type, although in any situation involving a finite number of entities, all but finitely many would have value *Absurd*.

Typed MEBN can represent uncertainty about the type of an entity, refine type-specific probability distributions through Bayesian learning, inherit distributions from parent types, and incorporate other features related to representing and reasoning with incomplete and/or uncertain information in typed systems (Costa, 2005). As an example, we might consider two subtypes of starships, fighters and cargo ships. When we are unsure about a starship's type, the result of a query that depends on type will be a weighted average of the result given that the ship is a fighter and the result given that it is a cargo ship.

Hierarchical Bayesian methods increase efficiency of parameter learning by allowing types with sparse data to "borrow strength" from related types with ample data. For example, the distribution for starship length might have a type-specific average length, which might depend probabilistically on features such as the purpose (e.g., warfare, commerce) or the species of the designer and manufacturer. Data relevant to the length distribution of one type of starship will help to refine the length distribution for similar types of starship by refining estimates of parameters related to features that influence length.

MEBN logic can also represent and reason about hypothetical entities. Uncertainty about whether a hypothesized entity actually exists is called *existence uncertainty*. In our example MTheory, the RV *Exists*($st$) represents whether or not its argument is an actual starship. For example, we might be unsure whether a sensor report corresponds to one of the starships we already know about, a starship of which we were previously unaware, or a spuri-ous report. To allow for hypothetical starships, the local distribution for *Exists*(*st*) assigns non-zero probability to *False*. Suppose the unique identifier !*ST*4 refers to a hypothetical starship nominated to explain the report. In this case, *Isa*(*Starship*, !*ST*4) has value *True*, but the value of *Exists*(!*ST*4) is uncertain. A value of *False* would mean !*ST*4 is a *spurious* starship or *false alarm*. Queries involving the unique identifier of a hypothetical starship return results weighted by our belief that it is an actual or a spurious starship. Belief in *Exists*(!*S*4) is updated by Bayesian conditioning as relevant evidence accrues. Representing existence uncertainty is especially useful for counterfactual reasoning and reasoning about causality (e.g., Pearl 2000).

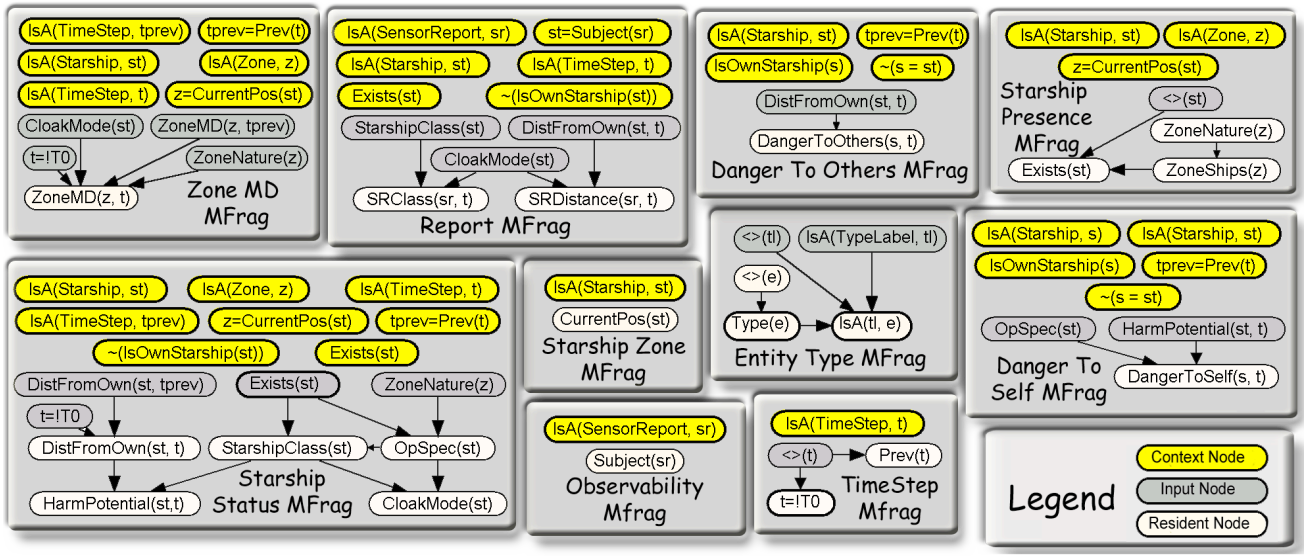

Figure 2 – The *Star Trek* Generative MTheory

MEBN logic can also represent *association uncertainty*, a major problem for multi-source fusion systems. Association uncertainty means we are not sure about the source of a given report. For example, a report (say, !*SR*4) may indicate a starship near a given location, but it may be unclear whether the report was generated by !*ST*1 or !*ST*3, two starships known to be near the reported location, or by a previously unreported starship !*ST*4. In this case, we would enumerate these three unique identifiers as possible values for *Subject*(!*SR*4), and specify that *Exists*(!*ST*4) has value *False* if *Subject*(!*SR*4) has any value other than !*ST*4. Many weakly discriminatory reports coming from possibly many starships produces an exponential set of combinations that require special *hypothesis management* methods (c.f. Stone *et al.* 1999). For example, we might not nominate !*ST*3 as a possible value for *Subject*(!*SR*4) if its distance from the reported location exceeded our *gating threshold*, even though if is logically possible for the report to have been generated by !*ST*3.

Closely related to association uncertainty is *identity uncertainty*, or uncertainty about whether two expressions refer to the same entity. Association uncertainty can be regarded as a special case of identity uncertainty – that is, we are unsure about the identity of *Subject*(!*SR*4). The ability to represent existence, association, and identity uncertainty provides a logical foundation for hypothesis management in multi-source fusion.

Finally, another key aspect of MEBN logic is its flexibility. The generative MTheory of Figure 2 is just one of many MTheories that could represent the same joint distribution. We grouped the RVs in a way that naturally reflects the structure of the objects in our scenario (i.e. we adopted an object oriented approach to modeling), but this was only one design option among many. Ultimately, the approach to be taken when building an MTheory depends on many factors, including the purpose, the background and preferences of the stakeholders, the need to interface with external systems, etc.

## 4 INFERENCE IN MEBN LOGIC

MEBN inference responds to queries for the degree of belief in target RVs given evidence RVs. We start with a generative MTheory, add a set of finding MFrags representing problem-specific information, and specify the target nodes for our query. We can compute the response to a query by constructing a *situation-specific Bayesian network* (SSBN). This is an ordinary Bayesian network constructed by combining instances of the MFrags in the generative MTheory. An standard Bayesian network inference algorithm is applied to the SSBN to answer the query.

Figure 3 shows an example of a SSBN constructed in response to findings asserting the presence of five starships (!*ST*0 to !*ST*4), the first being our own starship, plus data regarding the nature of the space zone (!*Z*0), its respective magnetic disturbance for the first time step (!*T*0), and sensor reports for starships !*SR*1 to !*SR*4 at !*T*0. The target set for this illustrative query includes an assessment of the level of danger experienced by the *Enterprise*.

In some cases the SSBN can be infinite, but under conditions given in Definition 2 above, the algorithm produces a sequence of approximate SSBNs for which the posterior distribution of the target nodes converges to their posterior distribution given the findings. Mahoney and Laskey (1998) define a SSBN as a minimal Bayesian network sufficient to compute the response to a query. A SSBN may contain any number of instances of each MFrag, depending on the number of entities and their interrelationships.

For a detailed account of the SSBN construction algorithm, the interested reader should refer to Laskey (2005). The paper covers a number of topics not treated here, such as nodes with an infinite number of states, countably infinite recursions, nodes with infinitely many parents, what happens when SSBN construction is applied to an inconsistent MTheory, etc. The paper also provides details on how to represent any FOL sentence as an MFrag, as well as an overview of Bayesian learning, which is treated in MEBN logic as a sequence of MTheories.

## 5 RELATED RESEARCH

Hidden Markov models, or HMMs (Baum & Petrie 1966), have been applied extensively in pattern recognition applications. HMMs can be viewed as a special case of dynamic Bayesian networks, or DBNs (Murphy 1998). A HMM is a DBN having hidden states with no internal structure that *d*-separate observations at different time steps. Partially dynamic Bayesian networks, also called temporal Bayesian networks (Takikawa et al. 2001), extend DBNs to include static variables. These formalisms augment standard Bayesian networks with a capability for temporal recursion.

Like MEBN logic, Bayesian logic programs (e.g., Kersting & De Raedt 2001) express uncertainty over models of first-order theories. Thus, their semantic basis is the same as MEBN logic. Bayesian logic programs and MTheories represent complementary approaches to specifying first-order probabilistic theories. BLPs represent fragments of

Bayesian networks in first-order logic; MEBN theories represent FOL sentences as MFrags.

BUGS (Gilks *et al.* 1994) is a software package based on plates. Plates represent repeated fragments of directed or undirected graphical models. Visually, a plate is represented as a rectangle enclosing a set of repeated nodes. Strengths of BUGS are the ability to handle continuous distributions without resorting to discretization, and support for parameter learning in a wide variety of parameterized statistical models. Object-oriented Bayesian Networks (Koller & Pfeffer 1997) represent entities as instances of object classes with class-specific attributes and probability distributions. Probabilistic Relational Models (PRMs) (Getoor *et al.* 2000) integrate the relational data model and BNs.

Results similar to Theorem 1 exist for both Bayesian logic programs and probabilistic relational models (e.g., Kersting & De Raedt 2001; Jaeger 1998). In particular, joint distributions can be specified over infinitely many random variables. Results for probabilistic relational models assume all random variables are binary, and in infinite domains only non-recursive models are allowed. Kersting and De Raedt's results for Bayesian logic programs are more general, allowing recursion and non-Boolean random variables. To our knowledge, Laskey (2005) is the first to demonstrate existence of a coherent joint distribution over models of any consistent, finitely axiomatizable first-order theory. This is a non-trivial extension to previous results, because in infinite domains, a non-contradictory axiom set may have zero probability, and there may be no well-defined conditional distribution. Because it is possible for any generative statistical theory to define an infinite sequence of findings to falsify the theory, no logic can express a coherent distribution over models of arbitrary infinite axiom sets (Laskey, 2005).

DAPER (Heckerman *et al.* 2004) combines the entity-relational model with DAG models to express probabilistic knowledge about structured entities and their relationships. Plate and PRM models can be represented in DAPER. Thus, DAPER is a unifying language for expressing relational probabilistic knowledge. DAPER expresses probability distributions over finite databases, and cannot represent arbitrary FOL expressions involving quantifiers. Therefore, like other languages discussed above, DAPER does not achieve full FOL representation power. MEBN provides the formal mathematical basis to achieve this objective, and the results of Laskey (2005) could be applied to extend the expressive power of any of the above formalisms.

In summary, as a full integration of first-order logic and probability, MEBN provides: (1) a means of expressing a globally consistent joint distribution over models of any consistent, finitely axiomatizable FOL theory; (2) a proof theory capable of identifying inconsistent theories in finitely many steps and converging to correct responses to probabilistic queries; and (3) a built in mechanism for adding sequences of new axioms and refining theories in the light of observations.

## 6 DISCUSSION AND FUTURE WORK

MEBN logic is a formal system that unifies probability theory and classical first-order logic. The ability to perform plausible reasoning with the expressiveness of FOL provides the potential to handle complex problems in a wide variety of application domains. The flexibility of the formal system defined in Laskey (2005) allows it to serve as the logical basis for any typed probabilistic knowledge representation language founded on FOL or a subset. For example, Quiddity*Suite™ (Fung *et al.* 2005) is a frame-based relational modeling toolkit that implements many features of MEBN logic and has been applied to a wide range of domains, including visual target recognition, multi-sensor fusion, cyber-security and our *Star Trek* scenario. Quiddity scripts for the *Star Trek* scenario can be found in Costa (2005).

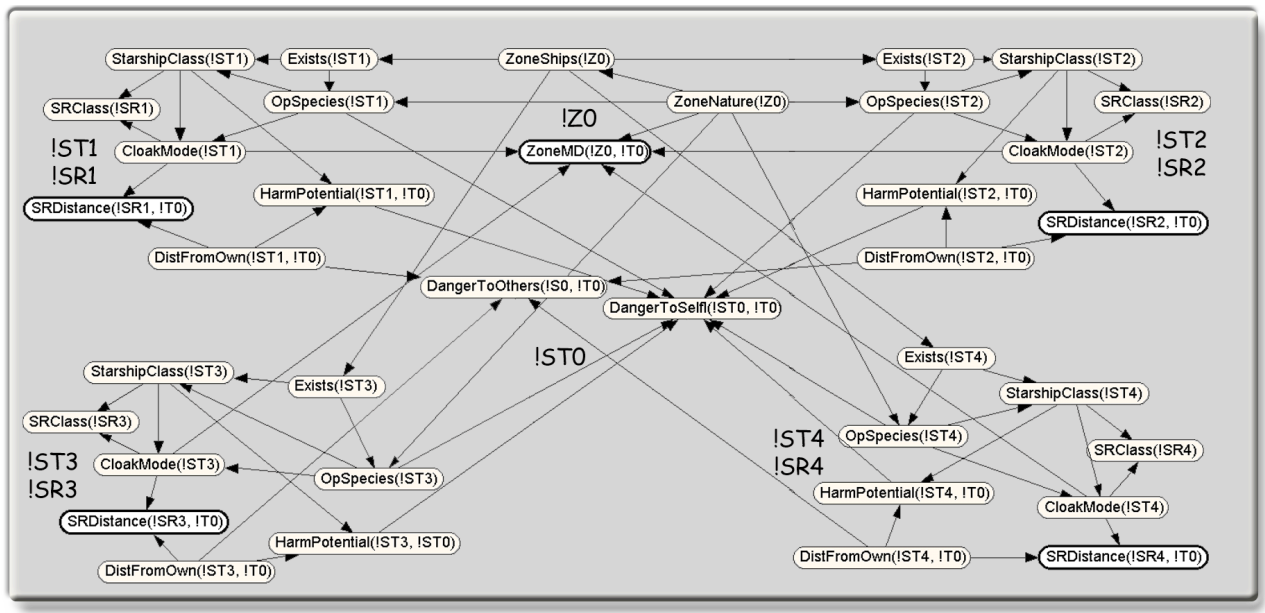

Figure 3 – A SSBN Built from the Star Trek Generative MTheory

XML-based languages such as RDF and OWL are currently being developed using subsets of FOL. MEBN can provide a logical foundation for extensions that support plausible reasoning. As an example, we are currently developing PR-OWL, a MEBN-based extension to the semantic web language OWL (Costa 2005). Our objective is to create a language capable of representing and reasoning with probabilistic ontologies. This technology has many possible applications to the Semantic Web, an open and highly uncertain environment for which expressive uncertain reasoning languages are sorely needed.

In summary, unification of Bayesian probability theory and classical first-order logic provides a formal basis for plausible reasoning in an open world characterized by many interacting entities related to each other in diverse ways and having many uncertain features and relationships.

## Acknowledgments

Grateful acknowledgement is due to the Brazilian Air Force for supporting Paulo Costa during his PhD studies at George Mason University. Kathryn Laskey's work was partially supported under a contract with the Office of Naval Research, number N00014-04-M-0277. The authors extend thanks to the GMU decision theory seminar participants whose many insightful questions helped us to clarify both our thinking and our writing, and to Francis Fung, Mike Pool, Masami Takikawa and Ed Wright for many helpful discussions. We thank the anonymous reviewers for thorough reviews that helped us to improve the paper. Last but not least, this paper is dedicated to Danny Pearl, and to the hope that in the 23[rd] Century, Danny's writings and his father's research will be remembered for their role in bringing about Danny's dream of a world in which all cultures and faiths live together in harmony.